\documentclass[11pt,a4paper]{article}
\usepackage[hyperref]{emnlp2018}
\usepackage{times}
\usepackage{latexsym}
\usepackage{CJK}
\usepackage{url}
\usepackage{natbib}
\usepackage{graphicx}
\usepackage{amsmath}
\usepackage{array}
\usepackage{multirow}
\usepackage{color,soul}

\aclfinalcopy 
\def\aclpaperid{1642} 

\newcommand*{\email}[1]{\texttt{#1}}

\title{ An Auto-Encoder Matching Model for Learning Utterance-Level Semantic Dependency in Dialogue Generation }

\author{
Liangchen Luo\thanks{\ \ Equal Contribution}, \
Jingjing Xu\footnotemark[1], \  
Junyang Lin, \ 
Qi Zeng, \ 
Xu Sun
\\ 
MOE Key Lab of Computational Linguistics, School of EECS, Peking University  \\
\email{\{luolc,jingjingxu,linjunyang,pkuzengqi,xusun\}@pku.edu.cn }}

\date{}

\begin{document}
\maketitle
\begin{CJK}{UTF8}{song}

\begin{abstract}

Generating semantically coherent responses is still a major challenge in dialogue generation. Different from conventional text generation tasks, the mapping between inputs and responses in conversations is more complicated, which highly demands the understanding of utterance-level semantic dependency, a relation between the whole meanings of inputs and outputs. To address this problem, we propose an Auto-Encoder Matching  (AEM) model to learn such  dependency. The model contains two auto-encoders and one mapping module. The auto-encoders learn the semantic representations of inputs and responses, and the mapping module learns to connect the utterance-level representations. Experimental results from automatic and human evaluations demonstrate that our model is capable of generating responses of high coherence and fluency compared to baseline models.\footnote{The code is available at \url{https://github.com/lancopku/AMM}} 




\end{abstract}

\section{Introduction}
Automatic dialogue generation task is of great importance to many applications, ranging from open-domain chatbots~\cite{DBLP:conf/coling/HigashinakaIMMKSHMM14, DBLP:journals/corr/VinyalsL15, DBLP:conf/emnlp/LiMRJGG16, DBLP:conf/emnlp/LiMSJRJ17,DBLP:conf/aaai/SuSHLC18} to goal-oriented technical support agents~\cite{DBLP:journals/corr/BordesW16, DBLP:journals/corr/abs-1712-02838, DBLP:conf/sigdial/AsriSSZHFMS17}. Recently there is an increasing amount of studies about purely data-driven dialogue models, which learn from large corpora of human conversations without hand-crafted rules or templates. Most of them are based on the sequence-to-sequence (Seq2Seq) framework~\cite{SutskeverVL14_seq2seq} that maximizes the probability of gold responses given the previous dialogue turn. Although such methods offer great promise for generating fluent responses, they still suffer from the poor semantic relevance between inputs and responses~\cite{jingjingxuemnlp18-01}. For example, given ``What's your name'' as the input, the models generate ``I like it'' as the output.

Recently, the neural attention mechanism~\cite{DBLP:conf/emnlp/LuongPM15,DBLP:conf/nips/VaswaniSPUJGKP17} has been proved successful in many tasks including neural machine translation~\cite{DBLP:journals/corr/abs-1805-04871} and abstractive summarization~\cite{DBLP:journals/corr/abs-1805-03989}, for its ability of capturing word-level dependency by associating a  generated word with relevant words in the source-side context. Recent studies~\cite{DBLP:conf/aaai/MeiBW17,DBLP:conf/aaai/SerbanKTTZBC17} have applied the attention mechanism to dialogue generation to improve the dialogue coherence. However, conversation generation is a much more complex and flexible task as there are less ``word-to-words'' relations between inputs and responses.
For example, given~\textit{``Try not to take on more than you can handle''} as the input and~\textit{``You are right''} as the response, each response word can not find any aligned words from the input. 
In fact, this task requires the model to understand the utterance-level dependency, a relation between the whole meanings of inputs and outputs.
Due to the lack of utterance-level semantic dependency, the conventional attention-based methods that simply capture the word-level dependency achieve less satisfying performance in generating high-quality responses.

To address this problem, we propose a novel \textbf{Auto-Encoder Matching} model to learn utterance-level dependency. First, motivated by ~\citet{DBLP:journals/corr/abs-1805-04869}, we use two auto-encoders to learn the semantic representations of inputs and responses in an unsupervised style. Second, given the utterance-level representations, the mapping module is taught to learn the utterance-level dependency. The advantage is that by explicitly separating representation learning and dependency learning, the model has a stronger modeling ability compared to traditional Seq2Seq models.   Experimental results show that our model substantially outperforms baseline methods in generating high-quality responses.

Our contributions are listed as follows:
\begin{itemize}
\item To promote coherence in dialogue generation, we propose a novel Auto-Encoder Matching model to learn the utterance-level dependency. 

\item In our proposed model, we explicitly separate utterance representation learning and dependency learning for a better expressive ability.

\item Experimental results on automatic evaluation and human evaluation show that our model can generate much more coherent text compared to baseline models.

\end{itemize}

\begin{figure}[t] 
\centering
\includegraphics[width = 0.8\linewidth]{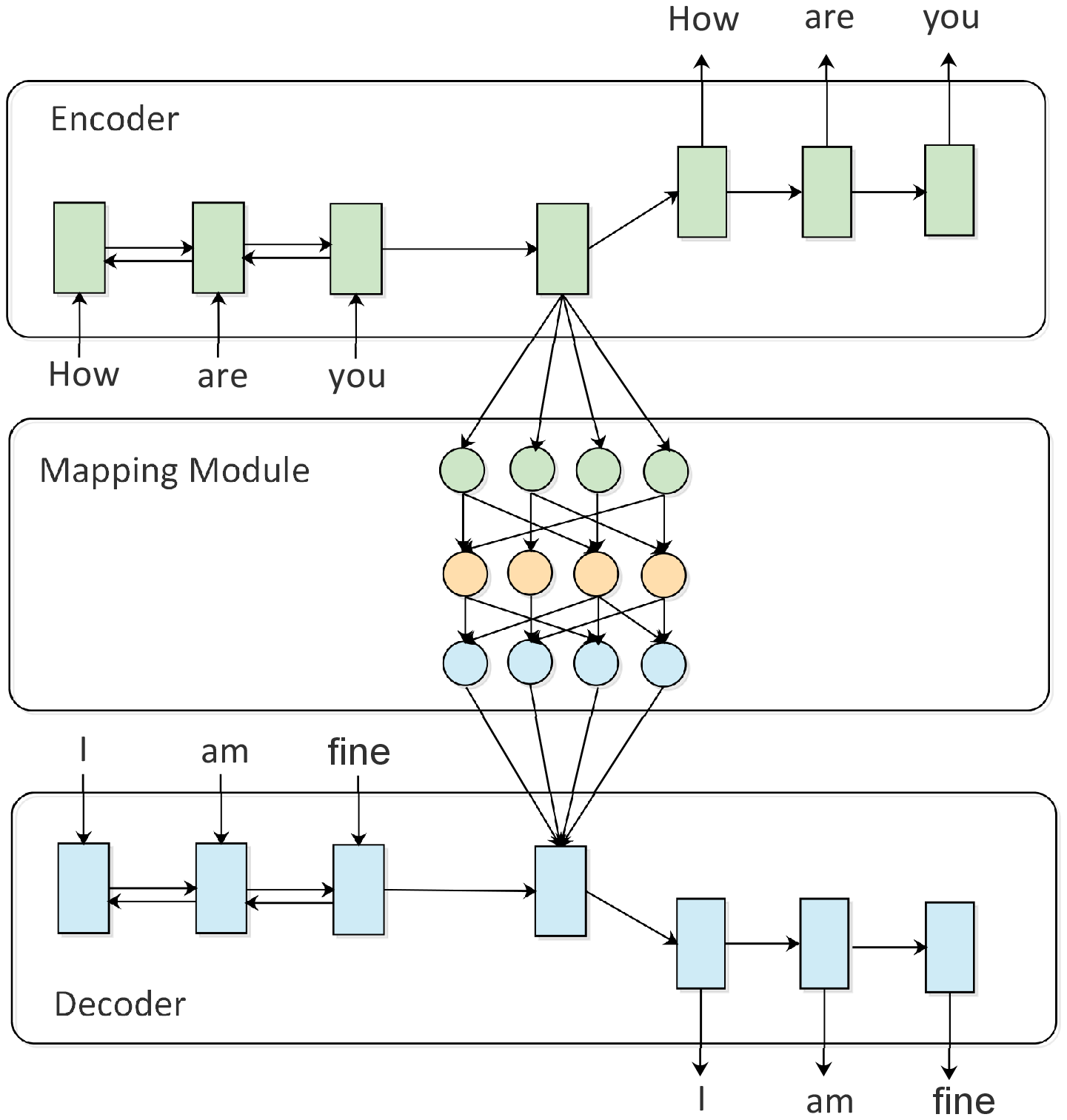}
\caption{An illustration of the Auto-Encoder Matching model. The encoder and decoder are two auto-encoders that are responsible for learning the semantic representations. The mapping module is responsible for learning the utterance-level dependency. }
\label{structure} 
\end{figure} 

\section{Approach}
In this section, we introduce our proposed model. An overview is presented in Section~\ref{overview}. The details of the modules are shown in Sections~\ref{encoder}, \ref{decoder} and \ref{mapping}. The training method is introduced in Section~\ref{training}.

\subsection{Overview}\label{overview}

The proposed model contains three modules: an encoder, a decoder, and a mapping module, as shown in Figure~\ref{structure}. 

In general, our model is different from the conventional sequence-to-sequence models. 
The encoder and decoder are both implemented as auto-encoders~\cite{DBLP:journals/jmlr/Baldi12}. They learn the internal representations of inputs and target responses, respectively.
In addition, a mapping module is built to map the internal representations of the input and the response.


\subsection{Encoder}\label{encoder}
The encoder $E_{\theta}$ is an unsupervised auto-encoder based on Long Short Term Memory Networks (LSTM) \citep{lstm}. As it is essentially a LSTM-based Seq2Seq model, we name the encoder and decoder of the auto-encoder ``source-encoder'' and ``source-decoder''. To be specific, the encoder $E_{\theta}$ receives the source text $x=\{x_1, x_2, ..., x_n\}$, and encodes it to an internal representation $h$, and then decodes $h$ to a new sequence $\tilde{x}=\{\tilde{x}_1, \tilde{x}_2, ..., \tilde{x}_n\}$ for the reconstruction of the input. We extract the hidden state $h$ as the semantic representation. The encoder $E_{\theta}$ is trained to reduce the reconstruction loss, whose loss function is defined as follows:
\begin{equation}
\begin{split}
J_1(\theta) =  -\log P(\tilde{x} |x; \theta)
\end{split}
\end{equation}
where $\theta$ refers to the parameters of the encoder $E_{\theta}$.


\subsection{Decoder}\label{decoder}
Similar to the encoder, our decoder $D_{\phi}$ is also a LSTM-based auto-encoder. However, as there is no target text provided in the testing stage, we propose the customized implementation, which is illustrated in Section~\ref{training}. Here in the introduction of the decoder, we do not provide the testing details. Similarly, we name the encoder and decoder of the auto-encoder ``target-encoder'' and ``target-decoder''. The target-encoder receives the target $y=\{y_1, y_2, ..., y_n\}$ and encodes it to a utterance-level semantic representation $s$, and then decodes $s$ to a new sequence to approximate the target text. The loss function is identical to that of the encoder:
\begin{equation}\label{learning-2}
\begin{split}
J_2(\phi) =  -\log P(\tilde{y} |y; \phi)
\end{split}
\end{equation}

\subsection{Mapping Module}\label{mapping}
As our model is constructed for dialogue generation, we design the mapping module to ensure that the generated response is semantically consistent with the source. There are many matching models that can be used to learn such dependency relations~\cite{DBLP:conf/nips/HuLLC14,DBLP:conf/cikm/GuoFAC16,DBLP:conf/aaai/PangLGXWC16,chendeli18}. For simplicity, we only use a simple feedforward network for implementation.  The mapping module $M_{\gamma}$ transforms the source semantic representation $h$ to a new representation $t$. To be specific, we implement a multi-layer perceptron (MLP) $g(\cdot)$ for $M_{\gamma}$ and train it by minimizing the L2-norm loss $J_3(\gamma)$ of the transformed representation $t$ and the semantic representation of target response $s$:
\begin{equation}
\begin{split}
t &= g(h)\\
J_3(\gamma) &= \frac{1}{2}\|t-s\|_2^2
\end{split}
\end{equation}

\subsection{Training and Testing}
\label{training}
In the testing stage, given an input utterance, the encoder $E_\theta$, the decoder $D_\phi$, and the matching module $M_\gamma$ work together to produce a dialogue response.
The source-encoder first receives the input $x$ and encodes it to a semantic representation $h$ of the source utterance. Then, the mapping module transforms $h$ to $t$, a target response representation. Finally, $t$ is sent to the target-decoder for response generation.

In the training stage, besides the auto-encoder loss and the mapping loss, we also use an end-to-end loss $J_4(\theta, \phi, \gamma)$ :
\begin{align}
	J_4(\theta, \phi, \gamma) &= - \log P(y|x;\theta, \phi, \gamma) \\
    &= - \sum_{t=1}^T \log P(y_t|x,y_{1..t-1};\theta, \phi, \gamma)
\end{align}

where $x$ is the source input, $y$ is the target response, and $T$ is the length of response sequence.
The model learns to generate $\tilde{y}$ to approximate $y$ by minimizing the reconstruction losses $J_1(\theta)$ and $J_2(\phi)$, the mapping loss $J_3(\gamma)$, and the end-to-end loss $J_4(\theta, \phi, \gamma)$. The details are illustrated below:
\begin{equation}
\begin{split}
J &= \lambda_1\left[J_1(\theta) + J_2(\phi)\right] + \lambda_2J_3(\gamma) \\
&+ \lambda_3J_4(\theta, \phi, \gamma)
\end{split}
\label{loss}
\end{equation}
where $J$ refers to the total loss, and $\lambda_1$, $\lambda_2$, and $\lambda_3$ are hyperparameters.

\section{Experiment}


We conduct experiments on a high-quality dialogue dataset called \textbf{DailyDialog} built by \citet{DBLP:conf/ijcnlp/LiSSLCN17}. The dialogues in the dataset reflect our daily communication and cover various topics about our daily life. We split the dataset into three parts with $36.3$K pairs for training, $11.1$K pairs for validation, and $11.1$K pairs for testing.




\subsection{Experimental Details}

For dialogue generation, we set the maximum length to 15 words for each generated sentence.
Based on the performance on the validation set, we set the hidden size to $512$, embedding size to $64$ and vocabulary size to $40$K for baseline models and the proposed model. The parameters are updated by the Adam algorithm~\cite{DBLP:journals/corr/KingmaB14} and initialized by sampling from the uniform distribution ($[-0.1, 0.1]$). The initial learning rate is $0.002$ and the model is trained in mini-batches with a batch size of $256$. $\lambda_1$ and $\lambda_3$ are set to $1$ and $\lambda_2$ is set to $0.01$ in Equation~\eqref{loss}. It is important to note that for a fair comparison, we re-implement the baseline models with the best settings on the validation set.
After fixing the hyperparameters, we combine the training and validation sets together as a larger training set to produce the final model.


\begin{table*}[t]
\setlength{\tabcolsep}{1pt}
\centering
    \begin{tabular}{l|c|c|c|c}
    \hline
    
   Models &BLEU-1 &BLEU-2 &BLEU-3 &BLEU-4 \\ \hline
    Seq2Seq & 12.43& 4.57 & 2.69&1.84 \\
    \textbf{AEM} & 13.55& 4.89 &3.04&2.16 \\ \hline
    Seq2Seq+Attention & 13.63 & 4.99 &3.05&2.13 \\
    \textbf{AEM+Attention} &\textbf{14.17} & \textbf{5.69} &\textbf{3.78}& \textbf{2.84}\\
    \hline
    \end{tabular}

    \caption{BLEU scores for the AEM model and the Seq2Seq model. }
    \label{tab:state}
        
\end{table*}

\begin{table*}[t]
\centering
    \begin{tabular}{l|c|c|c}
    \hline
  
 Models &  Dist-1  & Dist-2 & Dist-3 \\ \hline
  Seq2Seq  &0.8K & 2.7K & 5.5K               \\ 
    \textbf{AEM}  & 3.1K& 14.8K & 31.2K             \\ \hline
    Seq2Seq+Attention  & 2.5K& 13.6K & 34.6K             \\ 
     \textbf{AEM+Attention}  & \textbf{3.3K} & \textbf{23.2K} & \textbf{53.9K}             \\ 
  \hline

    \end{tabular}
    \caption{Diversity scores for the AEM model and the Seq2Seq model. Dist-1, Dist-2, and Dist-3 are respectively the number of distinct unigrams, bigrams, and trigrams in the generated text.  }
    \label{tab:auto-div}
\end{table*}

\subsection{Results}


\begin{table}[t]
\centering
\setlength{\tabcolsep}{2pt}
    \begin{tabular}{l|c|c|c}
    \hline
   
   
    Models & Fluency & Coherence & G-Score \\ \hline
    Seq2Seq& 6.97 & 3.51 & 4.95\\
   \textbf{AEM}   & \textbf{8.11} & 4.18 & 5.82\\\hline
   Seq2Seq+Attention& 5.11 & 3.30 & 4.10\\
   \textbf{AEM+Attention} & 7.92 & \textbf{4.97} & \textbf{6.27}\\\hline 
    \end{tabular}
    \caption{Human evaluation results of the AEM model and the Seq2Seq model. }
    \label{tab:human}
    
\end{table}

\begin{table}[t]
\centering
    \begin{tabular}{p{0.9\linewidth}}
    \hline
       \textbf{Input 1}: \textsl{My name is David. What is my name? }\\
        \textbf{Seq2Seq}: \textsl{ Ours, maria.}\\ 
    \textbf{AEM}: \textsl{ Pardon me? It is unreasonable!}\\ \hline
    
     \textbf{Input 2}: \textsl{I'm going to kill you! }\\
     \textbf{Seq2Seq}: \textsl{ I know. I'm a good man.}\\ 
    \textbf{AEM}: \textsl{ What?!?!!!! No!!}\\ \hline
    
     \textbf{Input 3}: \textsl{I hate you. }\\
     \textbf{Seq2Seq}: \textsl{I'm not. }\\
    \textbf{AEM}: \textsl{You do? you're a sweet girl.}\\ \hline

     \textbf{Input 4}: \textsl{How are you?}\\
     \textbf{Seq2Seq}: \textsl{ Fine.}\\ 
    \textbf{AEM}: \textsl{Fine. I'm good.}\\ \hline
    
      \textbf{Input 5}: \textsl{Can you say `Hello'? }\\
      \textbf{Seq2Seq}: \textsl{ Yeah, I guess.}\\ 
    \textbf{AEM}: \textsl{Sure, I guess.}\\ \hline
    
   

    \end{tabular}
    \caption{Examples generated by the AEM model and the Seq2Seq model. }
    \label{samplecases}
\end{table}



We use BLEU~\cite{DBLP:conf/acl/PapineniRWZ02}, to compare the performance of different models, and use the widely-used BLEU-4 as our main BLEU score. The results are shown in Table~\ref{tab:state}.
The proposed AEM model significantly outperforms the Seq2Seq model.
It demonstrates the effectiveness of utterance-level dependency on improving the quality of generated text. Furthermore, we find that the utterance-level dependency also benefits the learning of word-level dependency. 
The improvement from the AEM model to the AEM+Attention model\footnote{With the additional attention mechanism, the outputs of attention-based decoder and our decoder are combined together to predict the probability of response words.} is $0.68$ BLEU-4 point. It is much more obvious than the improvement from the Seq2Seq model to the Seq2Seq+Attention, which is $0.29$ BLEU-4 point. 





We also report the diversity of the generated responses by calculating the number of distinct unigrams, bigrams, and trigrams. The results are shown in Table~\ref{tab:auto-div}. We find that the AEM model achieves significant improvement on the diversity of generated text. The number of unique trigram of the AEM model is almost six times more than that of the Seq2Seq model. Also, it should be noticed that the attention mechanism performs almost the same compared to the AEM model ($31.2$K vs. $34.6$K in terms of Dist-3), which indicates that the utterance-level dependency and the word-level dependency are both indispensable for dialogue generation. Therefore, by combining the two dependencies together, the AEM+Attention model achieves the best results. Such improvements are expected. With the increase of the relevance of the generated text, it gets harder for the model to generate repeated responses. In our experimental results, the number of repetitive ``I don't know'' in the AEM+Attention model is reduced by $50$\% compared to the Seq2Seq model.

For dialogue generation, human evaluation is more convincing, so we also report human evaluation results on the test set. We randomly choose $100$ utterances in daily communication style for the human evaluation, each of which is sent to different models to generate responses. The results are distributed to the annotators who have no knowledge about which model the sentence is from. All annotators have linguistic background. They are asked to score the generated responses in terms of fluency and coherence. Fluency represents whether each sentence is in correct grammar. Coherence evaluates whether the generated response is relevant to the input. The score ranges from 1 to 10 (1 is very bad and 10 is very good). To evaluate the overall performance, we use the geometric mean of fluency and coherence as the final evaluation metric. 

Table~\ref{tab:human} shows the results of human evaluation. The inter-annotator agreement is satisfactory considering the difficulty of human evaluation. The Pearson's correlation coefficient is $0.69$ on coherence and $0.57$ on fluency, with $p<0.0001$. First, it is clear that the AEM model outperforms the Seq2Seq model with a large margin, which proves the effectiveness of the AEM model on generating high quality responses. Second, it is interesting to note that with the attention mechanism, the coherence is decreased slightly in the Seq2Seq model but increased significantly in the AEM model. It suggests that the utterance-level dependency greatly benefits the learning of word-level dependency. Therefore, it is expected that the AEM+Attention model achieves the best G-score.  



Table~\ref{samplecases} shows the examples generated by the AEM model and the Seq2Seq model. For easy questions (ex.~4 and ex.~5), they both perform well. For hard questions (ex.~1 and ex.~2), the proposed model obviously outperforms the Seq2Seq model. It shows that the utterance-level dependency learned by the proposed model is useful for handling complex inputs.

\subsection{Error Analysis}
Although our model achieves the best performance, there are still several failure cases. We find that the model performs badly for the inputs with unseen words. For instance, given ``Bonjour'' as the input, it generates ``Stay out of here'' as the output. It shows that the proposed model is sensitive to the unseen utterance representations. Therefore, we would like to explore more approaches to address this problem in the future work. For example, the auto-encoders can be replaced by variational auto-encoders to ensure that the distribution of utterance representations is normal, which has a better generalization ability.   

\section{Conclusion}
In this work, we propose an Auto-Encoder Matching model to learn the utterance-level semantic dependency, a critical dependency relation for generating coherent and fluent responses. The model contains two auto-encoders that learn the utterance representations in an unsupervised way, and a mapping module that builds the mapping between the input representation and response representation. Experimental results show that the proposed model significantly improves the quality of generated responses according to automatic evaluation and human evaluation, especially in coherence.  

\section*{Acknowledgements}

This work was supported in part by National Natural Science Foundation of China (No. 61673028). We thank all reviewers for providing the constructive suggestions. Xu Sun is the corresponding author of this paper.

%


\bibliography{emnlp2018}
\bibliographystyle{acl_natbib_nourl}

\end{CJK}
\end{document}